\documentclass[conference]{IEEEtran}
\IEEEoverridecommandlockouts
\usepackage{cite}
\usepackage{amsmath,amssymb,amsfonts}
\usepackage{algorithmic}
\usepackage{graphicx}
\usepackage{textcomp}
\usepackage{xcolor}
\usepackage{multirow}
\usepackage{enumitem} 
\usepackage[subrefformat=parens,labelformat=parens]{subfig}

\usepackage{wrapfig}

\renewcommand{\baselinestretch}{0.95} 

\def\BibTeX{{\rm B\kern-.05em{\sc i\kern-.025em b}\kern-.08em
    T\kern-.1667em\lower.7ex\hbox{E}\kern-.125emX}}
\IEEEaftertitletext{\vspace{-1.8\baselineskip}} %
\begin{document}

\title{{BPINN-EM-Post}: \underline{B}ayesian \underline{P}hysics-\underline{I}nformed \underline{N}eural \underline{N}etwork based Stochastic \underline{E}lectro\underline{m}igration Damage Analysis in the \underline{Post}-void Phase}
\vspace{-30pt}
\author{
    \IEEEauthorblockN{Subed Lamichhane,  Haotian Lu and Sheldon X.-D. Tan}
    \IEEEauthorblockA{Department of Electrical and Computer Engineering, University of California, Riverside, CA 92521}
    \IEEEauthorblockA{Email: slami002@ucr.edu, hlu123@ucr.edu, stan@ece.ucr.edu}
    \thanks{This work is supported in part by NSF grant under No.CCF-2007135 and in part by NSF grant CCF-2305437.}  
}
\maketitle

\begin{abstract}
\label{abstract}
In contrast to the assumptions of most existing Electromigration (EM) analysis tools, the evolution of EM-induced stress is inherently 
non-deterministic, influenced by factors such as input current fluctuations and manufacturing non-idealities. 
Traditional approaches for estimating stress variations typically involve computationally expensive and inefficient Monte Carlo simulations with industrial solvers, which quantify variations using mean and variance metrics. 
In this work, we introduce a novel machine learning-based framework, termed {\it \textbf{BPINN-EM-Post}}, for efficient stochastic analysis of EM-induced post-voiding aging processes. 
For the first time, our new approach integrates closed-form analytical solutions with a Bayesian Physics-Informed Neural Network (BPINN) framework to accelerate the analysis. 
The closed-form solutions enforce physical laws at the individual wire segment level, while the BPINN ensures that physics constraints at inter-segment junctions are satisfied and stochastic behaviors are accurately modeled. 
By reducing the number of variables in the loss functions through utilizing analytical solutions, our method significantly improves training efficiency without accuracy loss and naturally incorporates variational effects. 
Additionally, the analytical solutions effectively address the challenge of incorporating initial stress distributions in interconnect structures during post-void stress calculations. 
Numerical results demonstrate that {\it \textbf{BPINN-EM-Post}} achieves over $240\times$ and more than $67\times$ speedup compared to Monte Carlo simulations using the FEM-based COMSOL solver and FDM-based EMSpice, respectively, with marginal accuracy loss.

\end{abstract}

\section{Introduction}
\label{sec:Introduction}
Electromigration (EM) is a phenomenon where metal atoms within interconnects migrate due to interactions with high-current-carrying electrons, and considered as a major concern to VLSI interconnect. 
This atomic migration generates compressive stress at the anode and tensile stress at the cathode. 
When the stress exceeds a critical threshold, void formation near the cathode or hillock formation near the anode may occur, potentially causing circuit failure. 
The EM lifetime of interconnect wires is commonly characterized by the mean time to failure (MTTF), reflecting the inherent variability of the process~\cite{Li:ICCAD'11}.

Traditionally, the MTTF or reliability of an interconnect wire is linked to the wire’s current density through well-established EM models by Black and Blech~\cite{Black:1969fc, Blech:1976ko}. 
However, these single-wire segment EM models have been increasingly criticized for being overly conservative, as they treat each wire segment in isolation. 
Recent research demonstrates that stress evolution among wire segments within an EM tree, is highly correlated and should be analyzed collectively~\cite{Hauschildt:2013cv, Sukharev:2013tq}. 
Numerous physics-based analytical and numerical solutions have been proposed to address the above-mentioned limitations~\cite{DeORIO:Micro'2010, Sukharev:tdmr'16, ChenTan:TDMR'17, Chatterjee:2018TCAD, CookSun:TVSI'18, Abbasinasab:DAC'2018, ChenTan:TVLSI'19, SunYu:TDMR'20, Shohel:ICCAD'21, Lamichhane:SMACD'25}.
These methods primarily focus on solving Korhonen’s partial differential equation (PDE) to model EM-induced stress evolution~\cite{Korhonen:1993bb}. 
Nevertheless, traditional numerical techniques for solving Korhonen’s equation and related PDEs remain computationally costly. 
Furthermore, stochastic physics-based EM analyses using Monte Carlo simulations are prohibitively expensive for numerical solvers such as COMSOL~\cite{comsol:heat'2014} and EMSpice~\cite{SunYu:TDMR'20}, which demand multiple samplings to estimate variation distributions.

In recent years, machine learning–based methods have demonstrated strong effectiveness and efficiency in addressing complex EDA problems~\cite{SadiqbatchaZhao:ASPDAC'20, Jin:ICCAD'20, Lu:ICCAD'23, Lu:ISLPED'25}.
Among them, physics-informed neural networks (PINN) has been developed to facilitate the learning and integration of physical laws represented by nonlinear PDEs in complex physical, biological, and engineering systems~\cite{Raissi:JMLR'18, Raissi:JCP'19}. 
PINN approach has been utilized to solve Korhonen’s equation, showing promising results for EM stress modeling~\cite{HouChen:TCAD'22, Jin:ICCAD'22, HouZhen:TCAD'23}. 
In this framework, physical laws, boundary conditions (BCs), and initial conditions (ICs) of the PDEs are explicitly enforced through loss functions within the neural networks. 
This approach has demonstrated effectiveness on small-scale PDE problems.
Recently, hierarchical PINN schemes have been introduced to accelerate EM analysis~\cite{Jin:ICCAD'22, Lamichhane:ICCAD'23}.
These approaches divide the stress evolution problem in multi-segment interconnects into two stages, thereby significantly reducing the number of variables, as well as enhancing computational efficiency and scalability to large-scale interconnect structures. 
However, these PINN-based methods have yet to tackle the variability in EM effects.

Bayesian Neural Networks (BNNs) have emerged as an important class of neural networks due to their robust capabilities in uncertainty quantification, handling limited datasets, and preventing overfitting~\cite{Goan_2020}. 
Lamichhane \textit{et al.} introduced a hierarchical approach that integrates a pre-trained Bayesian network with a PINN framework to estimate EM stress variations, with a particular focus on the nucleation phase~\cite{Lamichhane:ICCAD'24}.
However, extending BNNs to analyze the post-voiding phase poses significant challenges, as it requires incorporating the initial stress distribution, a task that is highly complex and often infeasible for even simple neural networks~\cite{Lamichhane:ICCAD'23}. 

In this work, we address the above-mentioned challenges by leveraging analytical solutions for stress in single wire segments, ensuring accuracy through the enforcement of stress continuity and atomic flux conservation within the Bayesian PINN (BPINN) framework. 
Furthermore, the Bayesian component in BPINN facilitates a variational analysis of EM stress solutions.
Our major contributions are as follows:
\begin{itemize}[leftmargin=1em, labelsep=0.5em, parsep=2.5pt]
\item We propose a novel hierarchical PINN framework termed {\it \textbf{BPINN-EM-Post}} to address the stochastic stress evolution problem for EM damage assessment in multi-segment interconnect wires during the \textbf{post-voiding} phase. 
This framework, developed \textit{for the first time}, integrates analytical solutions with a BPINN for efficient training and accurate variational modeling.

\item Analytical solutions are utilized to precisely estimate the EM stress distribution within individual segments, regardless of void formulation. 
Unlike prior arts that rely on machine learning modeling, analytical approach overcomes the complexities associated with incorporating all possible initial stress distributions from nucleation phase into the post-voiding phase, enhances computational efficiency by reducing the number of variables within the framework.

\item The BPINN framework is specifically applied at the boundaries of single wire segments, working in conjunction with the analytical solutions. 
By using fewer variables in the loss function, the BPINN framework enforces key physical conditions, such as stress continuity and atomic flux conservation, while optimizing the atomic flux, which significantly accelerates the posterior sampling process.

\item Numerical results show that our proposed {\it \textbf{BPINN-EM-Post}} achieves a computational speedup of over $240\times$ compared to Monte Carlo simulations using FEM-based COMSOL, and more than $67 \times$ acceleration compared to Monte Carlo simulations using the FDM-based EMSpice method, with only negligible (less than $\sim 1\%$) error rate.

\end{itemize}

\section{Preliminaries}
\label{sec:prelim}
\subsection{EM Stress Evolution in Post-voiding Phase}
In confined metal wires subjected to high current densities, EM occurs as atoms are migrated from the cathode to the anode due to interactions between electrons and metal 
atoms~\cite{Black:1969fc}. 
Over time, this atom migration may potentially cause void and hillock formulations that compromise interconnect functionality.
Various methods have been developed to evaluate the EM-induced reliability of metal interconnects. 
Among these, Blech's limit~\cite{Blech:1976ko} and Black's MTTF~\cite{Black:1969fc} are traditional approaches for reliability assessment. 
Despite their utility, these methods are becoming less popular as Black's MTTF is limited to individual line segments and neither method accounts for the evolution of hydrostatic stress. Korhonen's equation~\cite{Korhonen:1993bb} addresses these limitations by describing EM in confined metal wires through a set of PDEs, and extends to model hydrostatic stress in multi-segment interconnects~\cite{WangYan:ICCAD'17,ChenTan:TVLSI'19}. 
The equation for the nucleation phase in multi-segment interconnects is given by Eq.~\eqref{eq:korhonen_multisegment_nucleation}, where $\sigma_{ij}(x,t)$ represents the stress in the interconnect segment $ij$ connecting nodes $i$ and $j$. 
In Eq.~\eqref{eq:korhonen_multisegment_nucleation}, $G_{ij}$ denotes the EM driving force in segment $ij$, calculated as $G_{ij} = \frac{e \rho J_{ij}Z^*}{\Omega}$, where $J_{ij}$ is the current density in segment $ij$.

\vspace{-10pt}
{\footnotesize
\begin{equation}
    \begin{aligned}
        PDE:& \frac{\partial \sigma_{ij}(x,t)}{\partial t}=\frac{\partial
        }{\partial x}\left[\kappa_{ij}(\frac{\partial \sigma_{ij}(x,t)}{\partial x} + G_{ij})\right],\ t>0  \\
        BC:& \sigma_{ij_1}(x_i,t)=\sigma_{ij_2}(x_i,t),\ t>0 \\
        BC:& \sum_{ij} \kappa_{ij}(\frac{\partial \sigma_{ij}(x,t)}{\partial x} \bigg |_{x=x_r}+G_{ij}) \cdot n_r = 0,\ t>0 \\
        BC:& \kappa_{ij}(\frac{\partial \sigma_{ij}(x,t)}{\partial x} \bigg |_{x=x_b}+G_{ij}) = 0,\ t>0 \\
        IC:& \sigma_{ij}(x,0)= \sigma_{ij,T}
        \label{eq:korhonen_multisegment_nucleation}
    \end{aligned}
\end{equation}}

The stress diffusivity denoted as $\kappa_{ij}$ is defined by $\kappa_{ij} = D_a B\Omega/(k_B T)$, where $D_a$ is the effective atomic diffusion coefficient, $B$ is the effective bulk modulus, 
$k_B$ is Boltzmann's constant, $T$ is the absolute temperature, and $E_a$ is the EM activation energy. 
$e$ is the electron charge, $\rho$ is resistivity, and $Z^*$ is the effective charge.

The first BC in Eq.~\eqref{eq:korhonen_multisegment_nucleation} enforces stress continuity at inter-segment junctions, specifically at $x = x_r$.
The second BC addresses atomic flux conservation at these junctions, while the third BC applies to blocking terminal boundaries $x = x_b$, ensuring zero atomic flux. 
The inward unit normal direction at the interior junction node $x_r$ on segment $ij$ is represented by $n_r$. 
The IC specifies that the initial stress distribution in segment $ij$ is given by $\sigma_{ij,T}$.

In multi-segment interconnect trees, void nucleation occurs at the cathode when the steady-state nucleation stress surpasses the critical stress $\sigma_{crit}$. 
This moment marks the nucleation time, denoted as $t_{nuc}$. 
Beyond this time, $t > t_{nuc}$, the void enlarges and the dynamics of stress evolution diverge from nucleation phase. 
Once nucleated, the stress at the interface of the void dramatically drops to zero~\cite{Sukharev:tdmr'16}.
For the segment where the void originates, the equations applicable in the post-voiding phase are outlined in Eq.~\eqref{eq:korhonen_equation_single_wire_postvoid}~\cite{Sukharev:tdmr'16}, which presumes the $x=0$ node as the cathode where the void forms. 
Should a void nucleate at $x=L$, the BCs are adjusted to reflect this scenario.
Eq.~\eqref{eq:korhonen_equation_single_wire_postvoid} posits that the void interface possesses an effective thickness, $\delta$, which is negligible relative to other dimensions within the interconnect tree. 
The examination of stress following void formation counts time, $t$, starting from the nucleation time, $t_{nuc}$. 
It is assumed that the initial stress distribution at this stage corresponds to the stress levels at the moment of void nucleation. 
This methodology facilitates the simulation of stress evolution within a single confined wire subsequent to the occurrence of void nucleation.

\vspace{-2pt}
{\footnotesize
\begin{equation}
    \begin{aligned}
        &PDE:  \frac{\partial \sigma(x,t)}{\partial t}=\frac{\partial
          }{\partial x}\left[\kappa(\frac{\partial \sigma(x,t)}{\partial x} + G)\right],\ t> 0  \\ 
        &BC: \frac{\partial \sigma(0,t)}{\partial x}\bigg |_{x=0}  = \frac{\sigma(0,t)}{\delta},\ t > 0 \\
        &BC: \frac{\partial \sigma(0,t)}{\partial x}\bigg |_{x=L} = -G,\ t > 0 \\ 
        &IC: \sigma (x,0) =  \sigma_{nuc}(x,t_{nuc}),\ t=0 \\ 
        \label{eq:korhonen_equation_single_wire_postvoid}
    \end{aligned}
\vspace{-10pt}
\end{equation}}

For analyzing multi-segment interconnects during the post-voiding phase, Eq.~\eqref{eq:korhonen_equation_single_wire_postvoid} is applied to the segment that contains the void. 
Segments without voids are addressed using Eq.~\eqref{eq:korhonen_multisegment_nucleation}. 
To accurately determine the stress profile across the entire interconnect tree, BCs that ensure stress continuity and atomic flux conservation must be implemented~\cite{Huang:TCAD'15}.

\subsection{Estimation of Variations in EM Stress Evolution}
Conventional deterministic approach solves the combinations of Eq.~\eqref{eq:korhonen_multisegment_nucleation} and Eq.~\eqref{eq:korhonen_equation_single_wire_postvoid} to determine the post-voiding phase transient stress distribution, which averages the DC input current, thus does not account for variations in EM stress caused by fluctuations in input current, manufacturing processes, temperature, and diffusivity. 
Consequently, this method tends to yield overly optimistic and unrealistic predictions of component lifetimes~\cite{Issa:ICCAD'20}.

If we define $\mathbf{x}$ as the vector of electrical and physical parameters influencing EM stress development, and $v$ as the error from these variations, the uncertainty and noise in these parameters can be treated as random variables within their respective spaces, i.e., $\mathbf{x} \in \mathbb{X}$ and $v\in \mathbb{V}$.  
It is posited that the solutions for EM stress are Gaussian distributed, centered around the true, but obscured, values~\cite{Issa:ICCAD'20} as shown in Eq.~\eqref{eq:variational_em_stress_model},

\vspace{-2pt}
{\footnotesize
\begin{equation}
    \begin{aligned}
        \bar{\sigma} = \mathcal{F}\left(\mathbf{x}\right) + v
    \end{aligned}
    \vspace{-2pt}
    \label{eq:variational_em_stress_model}
\end{equation}}
here $\bar{\sigma}$ represents the EM stress outcomes calculated using the parameter set $\mathbf{x}$ with the addition of noise $v$ from variability. 
$\mathcal{F}(.)$ could be any solver employed to solve combination of Eq.~\eqref{eq:korhonen_multisegment_nucleation} and Eq.~\eqref{eq:korhonen_equation_single_wire_postvoid}, such as the FEM-based COMSOL method~\cite{comsol:heat'2014}, or the FDM-based EMSpice tool~\cite{SunYu:TDMR'20}. 
Variation noise is typically assumed to be independent and normally distributed with a mean of zero and a specified variance $Var_{\sigma}$~\cite{Issa:ICCAD'20}.
A common technique for estimating the mean and variance in EM stress calculations, as described by combinations of Eq.~\eqref{eq:korhonen_multisegment_nucleation} and Eq.~\eqref{eq:korhonen_equation_single_wire_postvoid}, involves direct Monte Carlo simulation.
Nonetheless, Monte Carlo simulations necessitate repeated runs of combination of Eq.~\eqref{eq:korhonen_multisegment_nucleation} and Eq.~\eqref{eq:korhonen_equation_single_wire_postvoid}
for multi-segment interconnects, which can become exceedingly expensive, particularly as the number of segments in the interconnect structure grows.

The impact of variational or stochastic EM on the lifespan of wires has been explored in previous studies. 
\cite{Li:IRPS'14} highlights that as technology advances, both the MTTF and the distribution of failure times deteriorate, posing ongoing challenges for advanced interconnects. 
Early work~\cite{Li:ICCAD'11} introduced a method that incorporates variability into EM analysis. 
However, this approach examines a single wire segment individually and uses both analytical and numerical techniques to compute the time to failure. 
It considers two forms of variability: circuit-level variability, where the resistance of wires is treated as a random variable (RV), and wire geometry-level variability, where imperfections such as bumps and necking in wire geometries are RVs due to lithography-induced variations. 
~\cite{JainMishra:TVLSI'17} investigates both global and local process variations in power grids, applying a Hermite polynomial chaos method for variational EM analysis and final lifetime estimation using Black's EM model. 
In~\cite{Chatterjee:ICCAD'16}, a comprehensive analysis of the entire power grid is conducted, taking into account the inherent redundancy of grids by solving Korhonen's PDE using finite difference techniques. 
Here, EM diffusivity is treated as a RV, and Monte Carlo approach is employed for the variational analysis, supplemented by \textit{ad-hoc} acceleration techniques. 
\cite{Issa:ICCAD'20} primarily focuses on variability in input currents, where stochastic current models are used to model currents in functional blocks, and variance or covariance calculations are done via matrix-solving for entire power grids, which remains computationally expensive.
Work in \cite{Yang:JCP'21} introduced a Bayesian PINN framework aimed at quantifying noise in observations and addressing the overfitting challenges typical of standard PINNs, and showed effectiveness on standard, relatively straightforward problems. 
However, this method applies Bayesian-based sampling to the entire PINN during training, which can be quite time-consuming, especially for large-scale domains such as multi-segment interconnect structures.

Hence, in our proposed \textit{\textbf{BPINN-EM-Post}} framework, we combine fast and accurate analytical solutions with Bayesian PINN to efficiently estimate the variations in the EM stress on multi-segment interconnect structures. 
This hierarchical combination of the analytical solution and Bayesian PINN provides accurate estimation of variations in EM stress significantly faster than the conventional Monte Carlo simulations. 
In this work we follow the problem setting in~\cite{Issa:ICCAD'20} that considers the stochastic EM stress caused by fluctuations of input current density.
Other variations such as manufacture non-idealities can be modeled and analyzed through similar approach.
\section{Proposed BPINN-EM-Post Framework for Stochastic EM Analysis}
\label{sec:model}

\subsection{Estimation of Prior for BPINN}
\begin{figure}[h]
  \centering
  \vspace{-5pt}
  \includegraphics[width=\columnwidth]{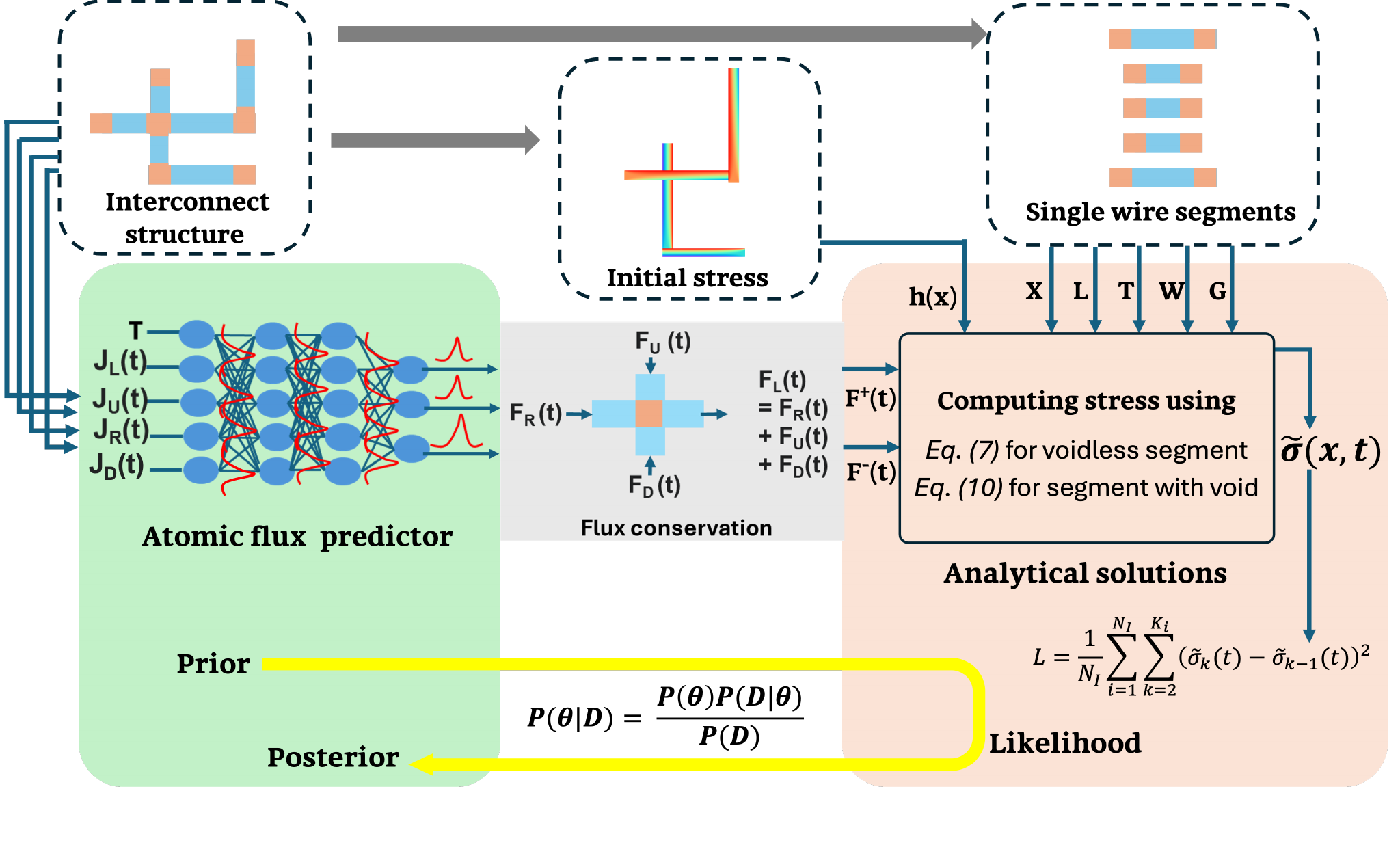}
  \vspace{-20pt}
  \caption{\small Framework of the proposed \textbf{\textit{BPINN-EM-Post}} variational EM simulator } 
  \vspace{-5pt}
  \label{fig:overall_framework}
\end{figure}

The overall framework of our proposed \textbf{\textit{BPINN-EM-Post}} is shown in Fig.~\ref{fig:overall_framework}. A fully connected network serves as a surrogate model for our Bayesian physics-informed neural network (BPINN) framework, based on prior work~\cite{Jin:ICCAD'22, Lamichhane:ICCAD'24} and the demonstrated capability of Multi-layer Perceptron (MLP) architecture for this prediction task.

Let each layer of the network be denoted as $l=1,2,3,\dots,L$ (where $L \ge 1$ is the number of layers). 
The output of the $l^{th}$ layer, $\mathbf{z}_l \in \mathbb{R}^{N_l}$, is expressed as Eq.~\eqref{eq:layer_output},

{\footnotesize
\begin{equation}
  \mathbf{z}_l = \phi\left(\mathbf{w}_{l-1} \mathbf{z}_{l-1} + \mathbf{b}_{l-1}\right)
  \label{eq:layer_output}
\end{equation}}
where $\phi$ is the non-linear activation function, $\mathbf{w} \in \mathbb{R}^{N_{l+1} \times N_l}$ represents weight matrices, and $\mathbf{b} \in \mathbb{R}^{N_{l+1}}$ denotes bias vectors. 
The network's output is given by Eq.~\eqref{eq:flux},

{\footnotesize
\begin{equation}
  \tilde{\mathbf{F}}(\mathbf{x};\omega) = \mathbf{w}_L \mathbf{z}_L + \mathbf{b}_L
  \label{eq:flux}
\end{equation}}
where $\mathbf{x}$ is the input vector. 

In this work, the Bayesian network processes information at all inter-segment junctions within a multi-segment interconnect structure. 
The input is $\mathbf{x} = \{T, J_L, J_U, J_R, J_D\}$, where $T$ is the temporal vector (time information) and $J_L, J_U, J_R, J_D$ are current densities for segments connected to the left, upper, right, and downward positions. 
The output $\tilde{\mathbf{F}}$ is 3-dimensional, with components $F_U$, $F_R$, and $F_D$ representing flux through the upper, right, and downward segments.

The unknown parameters $\omega$ comprise all weight matrices $\mathbf{w}$ and bias vectors $\mathbf{b}$. 
For the BNN, $\omega$ is assigned a prior such that its components are independent Gaussian distributions with mean $0$ and variances $\mathbf{Var}_{w,l}$ and $\mathbf{Var}_{b,l}$ for all layers~\cite{Yang:JCP'19}. 
As the hidden layer width approaches infinity, $\tilde{\mathbf{F}}(\mathbf{x})$ converges to a Gaussian process~\cite{Yang:JCP'19}.

\vspace{5pt}
\subsection{Generating Observations with Analytical Solutions}
 
To generate stress solutions $\tilde{\sigma}(x,t;\omega)$ using $\tilde{\mathbf{F}}(\mathbf{x};\omega)$, we adapt the analytical solutions proposed in~\cite{Lamichhane:ICCAD'23}. 
Before using the output from the Bayesian network, $\tilde{\mathbf{F}}(\mathbf{x};\omega)$, in the analytical solution, we perform linear transformations on it.

At any inter-segment junction of a multi-segment interconnect structure, up to four segments may connect. The three components of $\tilde{\mathbf{F}}$ represent flux information, which we denote as 
$F_U$, $F_R$, and $F_D$, corresponding to the upper, right, and downward segments as shown in Fig.~\ref{fig:overall_framework}. 
A linear transformation is applied to compute the flux through the remaining segment. 
For example, flux through the left segment is calculated as $F_L = F_U + F_R + F_D$, enforcing atomic flux conservation, as described in Eq.~\eqref{eq:korhonen_multisegment_nucleation}.

Using flux information at each inter-segment junction, we calculate flux values $F^-$ and $F^+$ at the endpoints of each segment within the multisegment interconnect structure. 
The analytical solutions in \textit{PostPINN-EM}~\cite{Lamichhane:ICCAD'23} use these endpoint flux values to compute stress on single wire segments. 
Here, $F^-$ and $F^+$ represent flux at the left/down and right/upper positions, respectively.

\begin{figure}[htp]
    \vspace{-20pt}
    \centering
    \subfloat[Segment without void.]{
        \includegraphics[width=0.45\columnwidth]{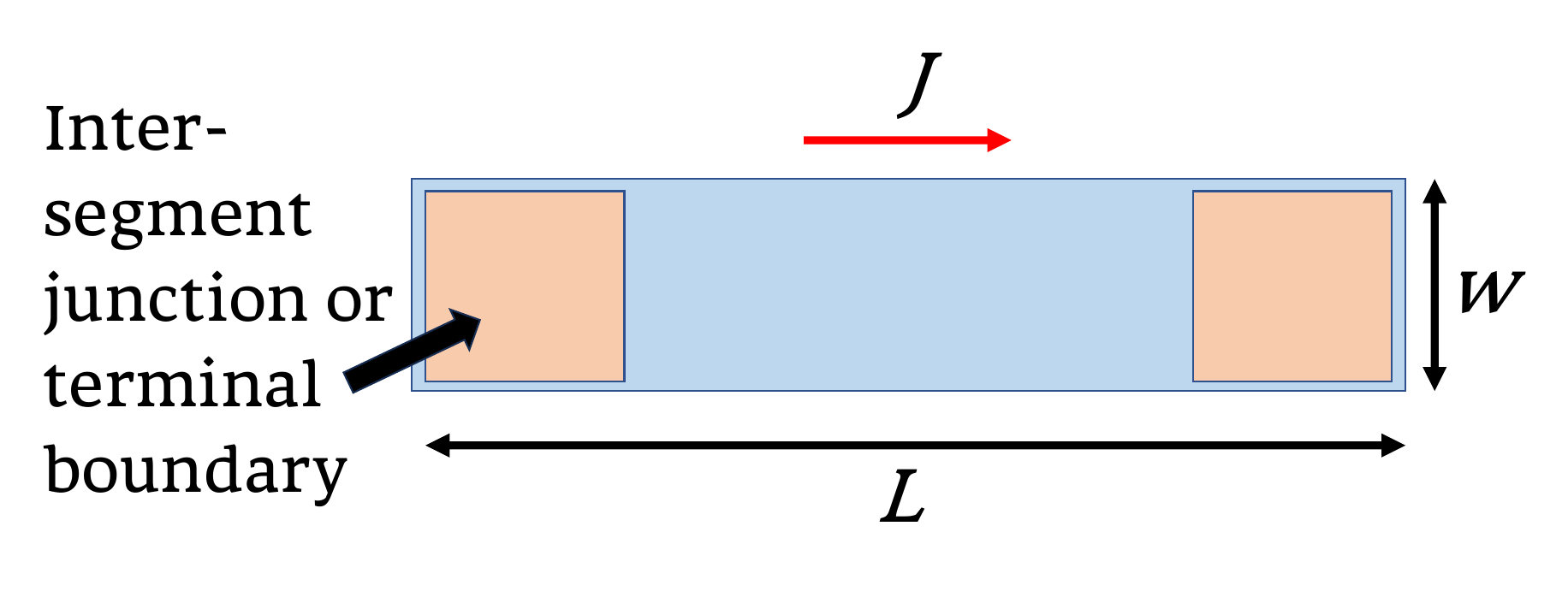}
        \label{fig:single_segment_voidless}
    }
    \hfill
    \subfloat[Segment with void at $x=L$.]{
        \includegraphics[width=0.45\columnwidth]{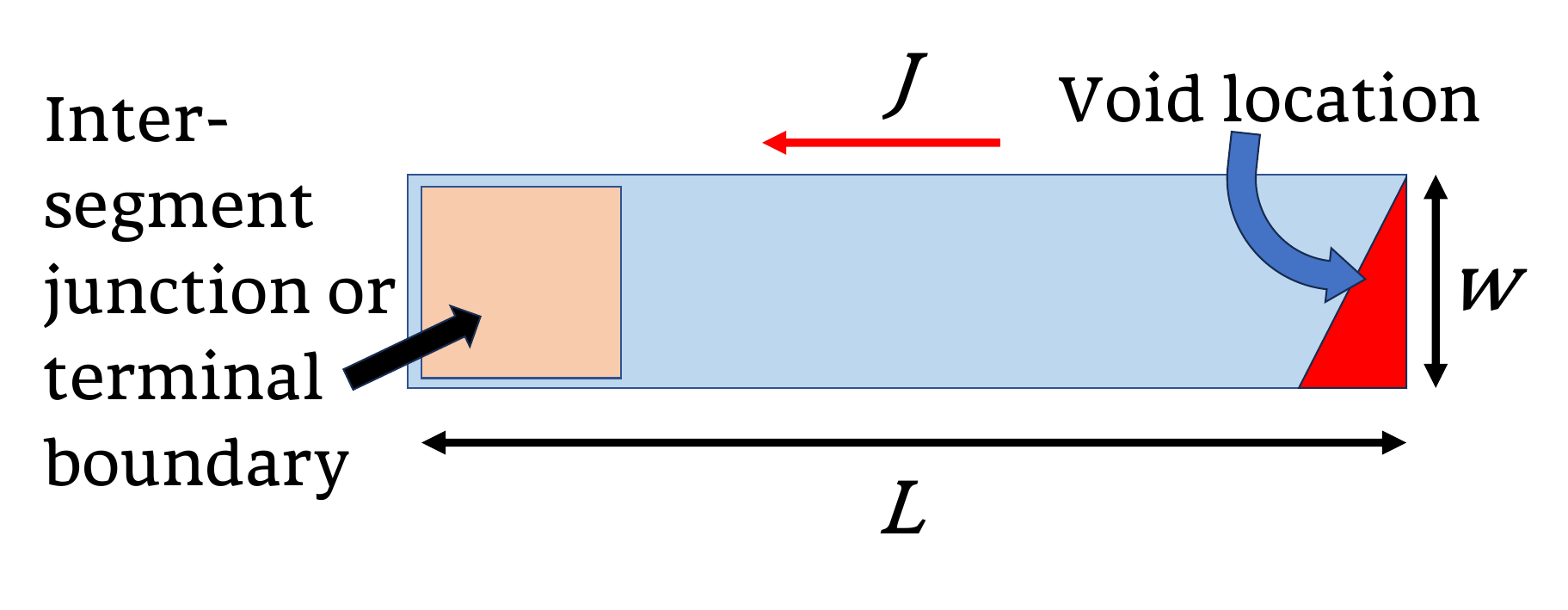}
        \label{fig:single_segment_void}
    }
    \vspace{-5pt}
    \caption{\small Single interconnect segments that are part of a multi-segment interconnect tree in the post-voiding phase.}
    \vspace{-5pt}
\end{figure}

As discussed in Sec.~\ref{sec:prelim}, two types of segments exist in the post-voiding phase of a multi-segment interconnect structure: a voidless segments as in Fig.~\ref{fig:single_segment_voidless}, and a segment containing a void at terminal as shown in Fig.~\ref{fig:single_segment_void}. 
Given $F^-$ and $F^+$, stress $\tilde{\sigma}(x,t)$ at any location and time on a single wire segment can be accurately calculated.

For analytical solutions, we define the stress gradient variables at the left or bottom node as $\phi^-(t)$ and at the right or top node as $\phi^+(t)$. 
Consequently, Korhonen's equation, modified with parametric BCs for wire segment without void is given in Eq.~\eqref{eq:korhonen_equation_single_wire_nucleation_varBC}, 

\vspace{-7pt}
{\footnotesize
\begin{equation}
  \begin{aligned}
      PDE&:\frac{\partial \tilde{\sigma}(x,t)}{\partial t}=\frac{\partial
        }{\partial x}\left[\kappa(\frac{\partial \tilde{\sigma}(x,t)}{\partial x} + G)\right],\ t>0,\ 0<x<L\\ 
      BC&: \frac{\partial \tilde{\sigma}(x,t)}{\partial x}\bigg |_{x = 0}  = \phi^-(t),\\
      BC&: \frac{\partial \tilde{\sigma}(x,t)}{\partial x}\bigg |_{x = L}  = \phi^+(t), \\
      IC&: \tilde{\sigma} (x,0) = \sigma_{nuc}(x,t_{nuc})=h(x)  
      \label{eq:korhonen_equation_single_wire_nucleation_varBC}
  \end{aligned}
\end{equation}}

In Eq.~\eqref{eq:korhonen_equation_single_wire_nucleation_varBC}, $\tilde{\sigma}(x,t)$ represents the revised stress response from the updated Korhonen's equation, while $\phi^-(t)$ and $\phi^+(t)$ serve as tunable BCs reflecting the stress gradients at the boundaries. 
It is assumed that $h(x)$ characterizes the initial stress state, corresponding to the stress at the nucleation moment $t_{nuc}$. 
The solution to Eq.~\eqref{eq:korhonen_equation_single_wire_nucleation_varBC} can be effectively derived using the Laplace transformation method, resulting in the following analytical solution for $\tilde{\sigma}(x,t)$ in the time domain, as shown in Eq.~\eqref{eq:voidless_segment_analytical_solution},

\vspace{-10pt}
{\footnotesize
\begin{equation}
\centering
  \begin{aligned}
    \tilde{\sigma}(x,t)
    &= \sum_{p=0}^{P}  \{ -\frac{d\phi^-(t)}{dt}*\left(g(\zeta_1(p,x,L),t)+ 
    g(\zeta_3(p,x,L),t) \right) \\
    &  - \phi^-(0) \left(g(\zeta_1(p,x,L),t)+ g(\zeta_3(p,x,L),t) \right) \\
    &  + \frac{d\phi^+}{dt} *\left(g(\zeta_2(p,x,L),t)+ g(\zeta_4(p,x,L),t) \right)  \\
    &  + \phi^+(0) \left(g(\zeta_2(p,x,L),t)+ g(\zeta_4(p,x,L),t) \right) \} \\
    &  +\frac{2}{L} \sum_{m=1}^M (-1)^m \{e^{-M}\cos(\frac{m\pi x}{L})\\
    & \times\int_0^L h(u) \cos \left(\frac{m\pi(L-u)}{L}\right) du \}
    \label{eq:voidless_segment_analytical_solution}
  \end{aligned}
\end{equation}
}

In this context, $p$ and $m$ are positive integers, with $0 \le p \le P$ and $1 \le m \le M$. 
It is recommended to maintain the three predominant components by setting $P=2$~\cite{HouZhen:TCAD'23}. 
The value of $M$ is established through experimental methods and is elaborately discussed in the results section. 
In Eq.~\eqref{eq:voidless_segment_analytical_solution}, the convolution operator is denoted by $\ast$, which is defined as $a(t) \ast b(t) = \int_{0}^{t} a(\tau) b(t-\tau) d\tau$. 
This convolution is efficiently executed using the Gauss-Legendre quadrature algorithm~\cite{HouZhen:TCAD'23}. 
The functions $\zeta_1(p,x,L),\ \zeta_2(p,x,L),\ \zeta_3(p,x,L)$, $\zeta_4(p,x,L),$ and $g(x,t)$ are specified in Eq.~\eqref{eq:zetas} as follows,

{\footnotesize
\begin{equation}
  \begin{aligned}
    & \zeta_1(p,x,L) = (2p+2)L - x,\ \\
    & \zeta_2(p,x,L) = (2p+1)L - x,  \\
    & \zeta_3(p,x,L) = (2p)L + x,\ \\
    & \zeta_4(p,x,L) = (2p+1)L + x, \\
    & g(x,t) = 2 \sqrt{\frac{kt}{\pi}}e^{-\frac{x^2}{4kt}} - 
    x \cdot \texttt{erfc} \{ \frac{x}{2\sqrt{kt}} \}
    \label{eq:zetas}
  \end{aligned}
\end{equation}}

For the segments with void, we assume that one end of the segment could either be an inter-segment junction or a terminal boundary, while the other end corresponds to a void node. Fig.~\ref{fig:single_segment_void} illustrates a single segment with a void at the $x=L$ end. 
At $x=0$, which may represent either an inter-segment junction or a terminal boundary, a modifiable BC from the voidless scenario is applied. 
We modify the BC at $x=L$ to $\tilde{\sigma}(L,t) = 0$. 
When using a very small $\delta$ in Eq.~\eqref{eq:korhonen_equation_single_wire_postvoid}, the BC at the site of the void mirrors the approach in~\cite{Bondurant_2005}. 
The BCs for the segment with a void at the subsequent node of a wire segment, i.e., at $x=L$, are expressed as Eq.~\eqref{eq:korhonen_equation_single_wire_postvoid_varBC},

{\footnotesize
\begin{equation}
  \begin{aligned}
      &BC: \frac{\partial \tilde{\sigma}(x,t)} {\partial x}\bigg |_{x=0}  = \phi^-(t) \\
      &BC: \tilde{\sigma}(x,t) |_{x=L} = 0
      \label{eq:korhonen_equation_single_wire_postvoid_varBC}
  \end{aligned}
\end{equation}}

Similar to the solution for the segment without a void, the solution for the single segment with a void location at $x=L$ is derived using the Laplace transform method as shown below in Eq.~\eqref{eq:void_segment_analytical_solution_plus},

\vspace{-10pt}
{\footnotesize
\begin{equation}
  \begin{aligned}
    \tilde{\sigma}(x,t)= &\sum_{p=0}^{P} (-1)^p \{\frac{d\phi^-(t)}{dt}*\left(g(\zeta_1(p,x,L),t)- 
                g(\zeta_3(p,x,L),t) \right) \\
                & +\phi^-(0) \left(g(\zeta_1(p,x,L),t)- g(\zeta_3(p,x,L),t) \right)\} \\
                & +\frac{2}{L} \sum_{m=1}^M (-1)^{m+1} \{e^{-M} \cos(\frac{(m-0.5)\pi x}{L})\\
                & \times \int_0^L h(u) \sin \left(\frac{(m-0.5)\pi(L-u)}{L}\right) du \}
    \label{eq:void_segment_analytical_solution_plus}
  \end{aligned}
\end{equation}}

Similarly, when the void is positioned at the preceding (left/bottom) end, i.e., at $x=0$, the BC is defined as $\tilde{\sigma}(x,t) = 0$ at $x=0$ and $\frac{\partial \tilde{\sigma}(x,t)}{\partial x} = \phi^+(t)$ at $x=L$. 
Using the same Laplace transform method, the analytical solution for this scenario can also be derived.
In Eq.~\eqref{eq:voidless_segment_analytical_solution} and Eq.~\eqref{eq:void_segment_analytical_solution_plus} the values $\phi^-(0)$ and $\phi^+(0)$ can be calculated using the current density information at each inter-segment junctions~\cite{Lamichhane:ICCAD'23, HouZhen:TCAD'23}.
For each single segment, we use the flux information from Bayesian network as $\frac{d\phi^-(t)}{dt} = F^-(\mathbf{x};\omega)$ and $\frac{d\phi^+(t)}{dt} = F^+(\mathbf{x};\omega)$. 

The analytical solutions in Eq.~\eqref{eq:voidless_segment_analytical_solution} and Eq.~\eqref{eq:void_segment_analytical_solution_plus} estimate the stress distribution for single segments within a multi-segment interconnect structure, given the flux information.
For accurate stress distribution, all relevant physical conditions must be satisfied. 
While calculating the flux information, atomic flux conservation was enforced. 
Additionally, as per Eq.~\eqref{eq:korhonen_multisegment_nucleation}, stress continuity must be maintained at inter-segment junctions. 
This implies that the stress in all segments connected to a junction should be equal.

To ensure stress continuity, we optimize the BPINN using the stress continuity condition $\mathcal{L}$, which is defined in Eq.~\eqref{eq:loss_function},

{\footnotesize
\begin{equation}
  \mathcal{L} = \frac{1}{N_{I} \times K_{i}} \sum_{i=1}^{N_{I}}
  \sum_{k=2}^{K_{i}} \left(\tilde{\sigma}_{k}(x_{\sigma};\omega) - \tilde{\sigma}_{k-1}(x_{\sigma};\omega)\right)^{2}
  \label{eq:loss_function}
\end{equation}}

where $\tilde{\sigma}_k(\omega)$ represents the EM stress at the boundary of the $k^{th}$ segment at inter-segment junction $i$, $K_i$ is the number of segments connected to 
junction $i$, and $N_I$ is the total number of inter-segment junctions in the multi-segment structure. 
Since $\tilde{\sigma}$ depends on $F^-(\mathbf{x};\omega)$ and $F^+(\mathbf{x};\omega)$, it is expressed as a function of $\omega$ and the parameters influencing stress, denoted as $x_{\sigma}$, for simplicity.

\vspace{5pt}
\subsection{Evaluation of Posterior for BPINN}

In this work, the Bayesian network takes time and current information at each inter-segment junction of the multi-segment interconnect structure as input, i.e., $\mathbf{x} = \{T, J_L, J_U, J_R, J_D\}$. Using the flux information from the network's output, we enforce atomic flux conservation to calculate the flux values $F^-(\mathbf{x};\omega)$ and $F^+(\mathbf{x};\omega)$ at the endpoints of each segment.

The analytical solutions use these flux values, along with geometric and electrical properties of the wires, to compute the stress $\tilde{\sigma}(x,t)$. 
To ensure accurate stress predictions, we incorporate stress observations at the boundaries of each wire segment to construct a loss function $\mathcal{L}$. 
This function is used to optimize the Bayesian network's parameters, yielding flux information that satisfies both atomic flux conservation and stress continuity at wire segment boundaries.

Therefore, for this case we can define the dataset as Eq.~\eqref{eq:dataset},

{\footnotesize
\begin{equation}
  \mathcal{D} = \left\{\mathbf{x}^{(i)},\bar{\mathcal{L}^{(i)}}\right\}_{i=1}^{N_{\mathcal{L}}}
  \label{eq:dataset}
\end{equation}}
here we assume that the observations are variational, and can be represented as in Eq.~\eqref{eq:loss},

{\footnotesize
\begin{equation}
  \bar{\mathcal{L}} = \mathcal{L}(x_{\mathcal{L}}^{(i)}) + \epsilon^{(i)}, i = 1,2,3,..N^{\mathcal{L}}
  \label{eq:loss}
\end{equation}}
where $\epsilon$ is independent Gaussian noise with zero mean and variance of $\mathbf{Var_{\mathcal{L}}}$. 
The likelihood of the observations can be calculated as Eq.~\eqref{eq:likelihood},

\vspace{-10pt}
{\footnotesize
\begin{equation}
\centering
  \begin{aligned}
      P(\mathcal{D}|\omega) = & \prod_{i=1}^{N_{\mathcal{L}}} \frac{1}{\sqrt{2\pi}Var_{\mathcal{L}}^{(i)}} \times exp \left[-\frac{\left(\mathcal{L}\left(x_{\mathcal{L}}^{(i)};w\right)-\bar{\mathcal{L}}^{(i)}\right)^2}{2Var_{\mathcal{L}}^{(i)}}\right]
  \end{aligned}
  \label{eq:likelihood}
\end{equation}}
Utilizing this likelihood, the posterior can be obtained as,

{\footnotesize
\begin{equation}
  \begin{aligned}
      P(\omega|\mathcal{D}) = \frac{P(\mathcal{D}|\omega)P(\omega)}{P(\mathcal{D})} \propto P(\mathcal{D}|\omega)P(\omega)
  \end{aligned}
  \label{eq:bayes_theorem}
\end{equation}}
here the symbol $\propto$ signifies 'equality up to a constant'. 
The probability of the dataset $P(D)$ is not readily solvable through analytical methods. 
Therefore, in practical scenarios, we only obtain an un-normalized expression $P(\omega|\mathcal{D})$.

This work \textit{\textbf{BPINN-EM-Post}} employs the Hamiltonian Monte Carlo (HMC) method to navigate the parameter space of our BNN model. 
HMC is a gradient-based Markov Chain Monte Carlo (MCMC) technique that utilizes Hamiltonian dynamics for parameter exploration. 
Our HMC procedure begins by simulating Hamiltonian dynamics through a numerical integration method, followed by a correction using the Metropolis-Hastings acceptance step. 
Given a dataset $\mathcal{D}$, we postulate the target posterior distribution for $\omega$ as shown in Eq.~\eqref{eq:posterior},

{\footnotesize
\begin{equation}
  P(\omega|\mathcal{D}) \simeq exp(-U(\omega))
  \label{eq:posterior}
\end{equation}}
where $U(\omega) = -\log(P(\mathcal{D}|\omega)) - \log(P(\omega))$ serves as the potential energy in the system. 
To sample from the posterior distribution, HMC introduces an auxiliary momentum variable $r$, which is used to formulate a Hamiltonian system as,

{\footnotesize
\begin{equation}
  \begin{aligned}
    H(\omega,r) = U(\omega) + \frac{1}{2}r^T \mathrm{M}^{-1}r
  \end{aligned}
  \label{eq:hamilton}
\end{equation}}
Here, $M$ is a mass matrix usually set as identity matrix. 
HMC then generates samples from the joint distribution of $(\omega,r)$ as,

{\footnotesize
\begin{equation}
  \pi(\omega,r) \sim exp(-U(\omega) - \frac{1}{2} r^T \mathrm{M}^{-1}r)
\end{equation}}
where the samples of $\omega$ have a marginal distribution as we discard the samples of $r$. 
These samples are derived from the following Hamiltonian dynamics as in Eq.~\eqref{eq:hamiltonian_dynamics},

{\footnotesize
\begin{equation}
  \begin{aligned}
    d\omega = \mathrm{M}^{-1}rdt,\ \ dr = - \nabla U(\omega)dt
  \end{aligned}
  \label{eq:hamiltonian_dynamics}
\end{equation}}
Eq.~\eqref{eq:hamiltonian_dynamics} is discretized through the leapfrog method, and the Metropolis-Hastings step is employed to minimize the discretization error.
More implementation details on the HMC algorithm are explained in~\cite{Yang:JCP'19}.

\subsection{Estimation of Variations in EM Stress}

Using HMC, we sample the optimal Bayesian network parameters, $\omega$, to generate samples of flux information, $\small \left\{\tilde{\mathbf{F}}(\mathbf{x};\omega^{(i)})\right\}_{i=1}^M$. 
From these, we derive samples of $\{\mathbf{F}^-\}$ and $\{\mathbf{F}^+\}$. 
For each wire segment, these flux samples are used to compute $M$ stress solution samples, $\{\tilde{\sigma}(x,t)\}$.

Notably, when obtaining the optimal parameter samples via HMC, we utilize only stress observations at the boundaries of single wire segments within the multi-segment interconnect structure. 
This significantly reduces the variables in the loss function $\mathcal{L}$, enabling efficient training. 
Once the framework is fully trained and optimal parameters are obtained, we calculate the flux information at all wire segment boundaries. 
This flux information is termed optimal as it satisfies flux conservation and stress continuity at inter-segment junctions.
Using the optimal flux information samples, we compute accurate stress distribution samples for individual segments through analytical solutions. 
Additionally, variations in stress solutions, such as mean and standard deviation, are estimated from the calculated stress samples.
\section{Experimental Results}
\label{sec:ExperimentalResults}

\subsection{Experimental Setup}
The proposed \textit{\textbf{BPINN-EM-Post}} framework is fully developed using Python/PyTorch. 
The training and test procedures are conducted on a Linux server equipped with two Intel 22-core E5-2699 CPUs, 320 GB of memory, and an Nvidia TITAN RTX GPU.
A three-layer MLP with layer structure [I5-FC50-FC50-O3] is used in BPINN flux predictor.

\subsection{Data Preparation and Scaling}
To determine the stress distribution in a multi-segment interconnect structure, we first gather the physical and material attributes of the interconnect segments. 
Each segment $i$ is characterized by its physical properties, including length $L_i$ and width $w_i$, while assuming uniform thermal conductivity $\kappa$ across all segments. 
Spatial points $x$ are sampled within $[0, L]$, and temporal points are chosen from $t = 0$ to $1 \times 10^9$ s at intervals of $1 \times 10^7$ s. 
In this analysis, we incorporate the nucleation-phase solutions for these interconnects that are easily pre-established by existing methods~\cite{SunYu:TDMR'20, comsol:heat'2014, Lamichhane:ICCAD'24} consistent to this work, providing initial stress distribution $h(x)$ and void locations.

\begin{figure*}[htp]
    \centering
    \vspace{-30pt}
    \hspace{-36pt}%
    \subfloat[Single segment without void.]{
        \includegraphics[width=0.35\textwidth]{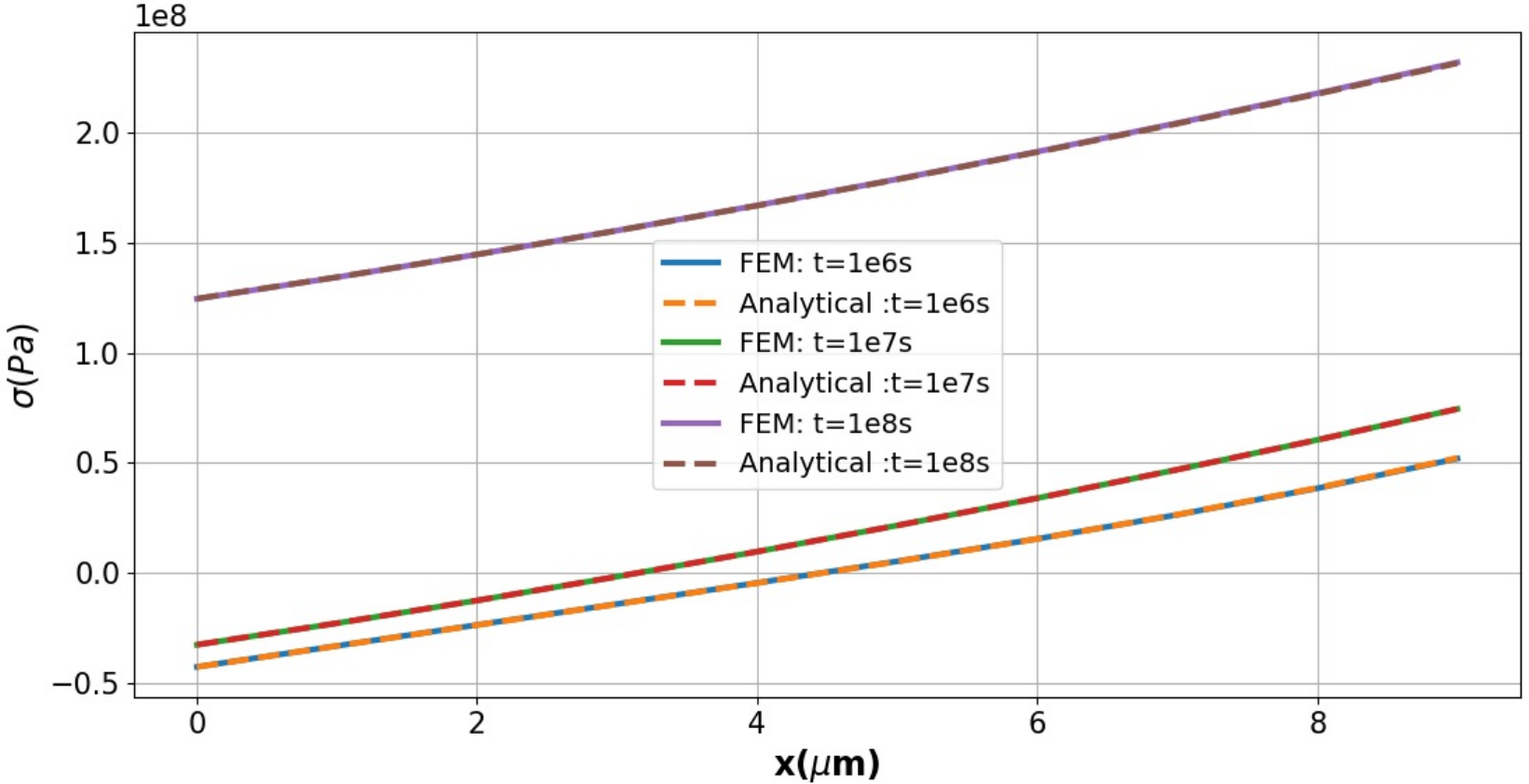}\hspace{-4pt}%
        \label{fig:case1}
    }
    \hfill
    \subfloat[
    Single segment with void at \textit{left} node.
    ]{
        \includegraphics[width=0.35\textwidth]{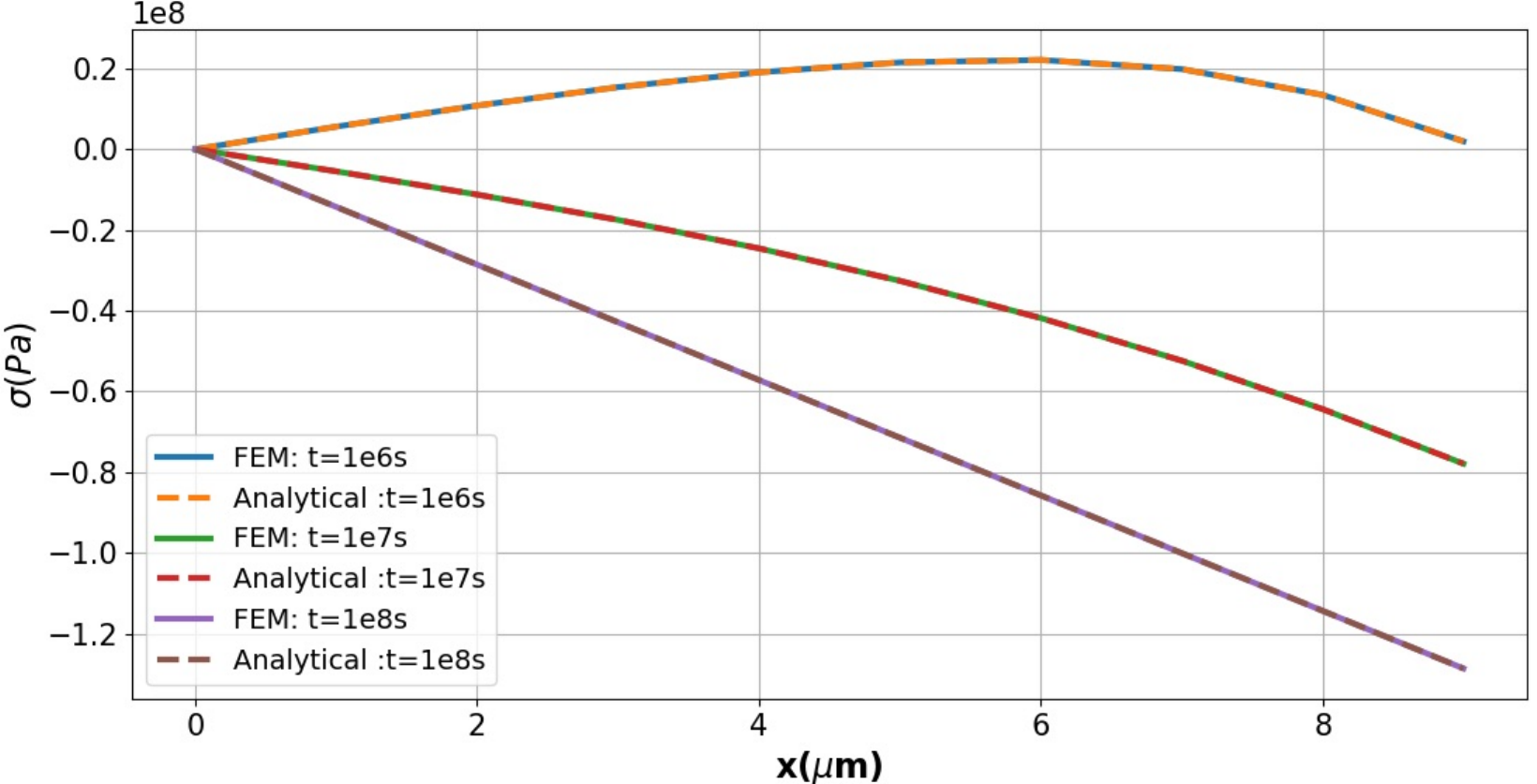}\hspace{-4pt}%
        \label{fig:case2}
    }
    \hfill
    \subfloat[Single segment with void at \textit{right} node.]{
        \includegraphics[width=0.35\textwidth]{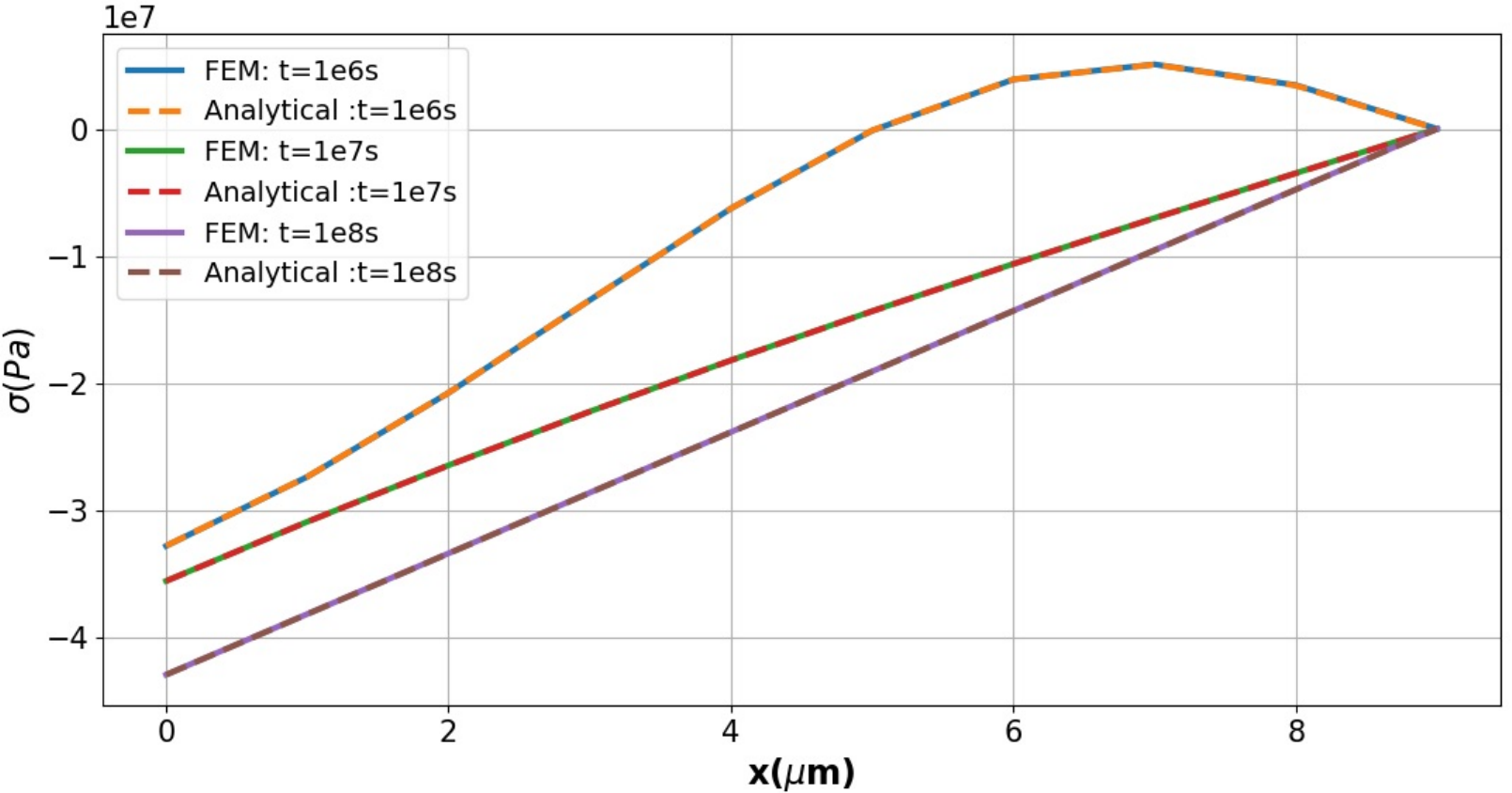}\hspace{-6pt}%
        \label{fig:case3}
    }
    \hfill
\vspace{-5pt}
\caption{
\small Comparison of the stress distribution obtained from analytical solution with the stress distribution from FEM-based COMSOL for single interconnect segment for three different cases.
}
\vspace{-5pt}
\label{fig:analytical_vs_comsol}
\end{figure*}

\renewcommand{\arraystretch}{1.3} %
\begin{table*}[!htp]
\centering
\caption{\small Performance and accuracy comparisons with existing methods~\cite{comsol:heat'2014},~\cite{SunYu:TDMR'20}}
\resizebox{\linewidth}{!}{
    \begin{tabular}{c|c|c|cccccc}
    \hline
    \multirow{2}{*}{\# of wires} & COMSOL~\cite{comsol:heat'2014}      & EMSpice~\cite{SunYu:TDMR'20}     & \multicolumn{6}{c}{The Proposed \textit{\textbf{BPINN-EM-Post}}}                                                                                                              \\ \cline{2-9} 
                                 & Runtime & Runtime  & Error v.s. COMSOL & Error v.s. EMSpice & Training time & Inference time & Speedup v.s. COMSOL & Speedup v.s. EMSpice \\ \hline
    50                           & 2560s        & 876s         & 0.25\%                   & 0.051\%                    & 6.5s                   & 0.52s                   & 364.67$\times \uparrow$             & 124.79$\times \uparrow$                \\
    100                          & 4800s        & 1490s        & 0.31\%                    & 0.074\%                    & 15.3s                  & 0.71s                   & 299.81$\times \uparrow$               & 93.07$\times \uparrow$                 \\
    150                          & 6731s        & 2122s        & 0.52\%                    & 0.094\%                    & 21.8s                  & 0.91s                   & 296.39$\times \uparrow$               & 93.44$\times \uparrow$                 \\
    200                          & 9230s        & 3100s        & 0.74\%                    & 0.126\%                    & 37.9s                  & 1.25s                   & 235.76$\times \uparrow$               & 79.18$\times \uparrow$                 \\
    250                          & 10191s      & 3656s        & 1.10\%                    & 0.163\%                    & 55.6s                  & 1.51s                   & 178.45$\times \uparrow$               & 64.02$\times \uparrow$                 \\ \hline
    \end{tabular}
}
\label{tab:result_statistics_line}
\vspace{-10pt}
\end{table*}

The input parameters for the analytical method and the MLP vary significantly in scale, necessitating data scaling for standardization. 
Scaling factors $k_x$, $k_t$, and $k_{\sigma}$ are used for spatial ($x$), temporal ($t$) and stress ($\sigma$), respectively. 
Additional parameters are processed in Eq.~\eqref{eq:scaling_scheme},

{\footnotesize
\begin{equation}
  \begin{aligned}
    k_{\sigma}\sigma(x,t,\kappa,G) &= \sigma_{sc}(x_{sc},t_{sc},\kappa_{sc},G_{sc})\\
    &= \sigma_{sc}(k_x x, k_t t, \frac{k_x^2}{k_t}\kappa, \frac{k_{\sigma}}{k_x}G)
  \end{aligned}
  \label{eq:scaling_scheme}
\end{equation}}
where $x_{sc}$, $t_{sc}$, $\kappa_{sc}$, $G_{sc}$, and $\sigma_{sc}$ are the scaled versions of $x$, $t$, $\kappa$, $G$, and $\sigma$, respectively. The scaled stress $\sigma_{sc}$ is converted back to its original scale using $k_{\sigma}$. 
In this study, scaling constants are set as $k_x = 1 \times 10^{-5}$, $k_t = 1 \times 10^{-7}$, and $k_{\sigma} = 1 \times 10^{-8}$.

To assess the accuracy and performance of our proposed \textit{\textbf{BPINN-EM-Post}}, we compare results against Monte Carlo simulations using FEM-based COMSOL~\cite{comsol:heat'2014} and FDM-based EMSpice tool~\cite{SunYu:TDMR'20}. 
For each Monte Carlo iteration, branch currents are randomly sampled from a normal distribution with a mean range of $5 \times 10^{-8} \, \mathrm{A/m^2}$ to $5 \times 10^9 \, \mathrm{A/m^2}$ to introduce variability to input currents. 

\subsection{Accuracy and Performance of Analytical Solutions}
In order to evaluate the precision and speed of the analytical solutions, we conducted comparisons with results from COMSOL simulations using a variety of specific stress gradients. 
We randomly generated 500 single wire segments samples, with lengths varying from $10\mu m$ to $50\mu m$, all maintaining a uniform width of $1 \mu m$.
These segments were subjected to current densities spanning from $-5\times10^9 A/m^2$ to $5\times10^{10} A/m^2$. 
Other physical parameters are set as $e = 1.6 \times 10^{-19} C$, $Z^* = 10$, $E_a = 1.1eV$, $B = 1\times10^{11}$, $D_0 = 5.2 \times 10^{-5}m^2/s$, $\rho = 2.2 \times 10^{-8} \Omega m$, $\Omega = 8.78 \times 10^{-30} m^3$, and $\sigma_{crit} = 5 \times 10^8 Pa$.
In the evaluation of our analytical solutions, the wire segments were modeled with non-zero boundary flux to represent their placement within the multi-segment trees. 
For each segment, we collected data at 30 spatial positions $x\in [0,L]$ and across 100 time intervals from 0 to $1\times10^{8}$ seconds to account for aging time $t$. 
These segments were individually analyzed using COMSOL and benchmarked against our analytical solution. 
We employed the root mean square error (RMSE) to measure discrepancies, noting that the stress values $\sigma(x,t)$ ranged from $-1.1\times10^{8}$ to $4.6\times10^{8}$ Pa. 
With our analytical approach, we achieved an \textbf{average RMSE of} $\mathbf{3.3\times10^{4}}$ Pa, correlating to an \textbf{average error margin of 0.006\%}.
The analytical solver recorded an average processing time of 0.01 seconds per segment, in stark contrast to COMSOL’s 7.5 seconds, resulting in a \textbf{computational speedup of approximately} $\mathbf{750\times}$. 
This demonstrates that our first-stage analytical solver delivers both rapid and precise results when juxtaposed with FEM-based COMSOL simulations.

Fig.~\ref{fig:analytical_vs_comsol} illustrates the comparative results of our analytical solution versus COMSOL for segments with and without voids, achieved with $M=5$ in the analytical calculations, which was decided by experimental determination and consistently applied throughout this study.

\subsection{Overall Accuracy and Performance Analysis} 

\begin{figure}[htp]
    \centering
    \subfloat[Ten-segment interconnect structure. Depicted current densities are the mean values of the current densities used during simulation.]{
        \includegraphics[width=0.85\linewidth]{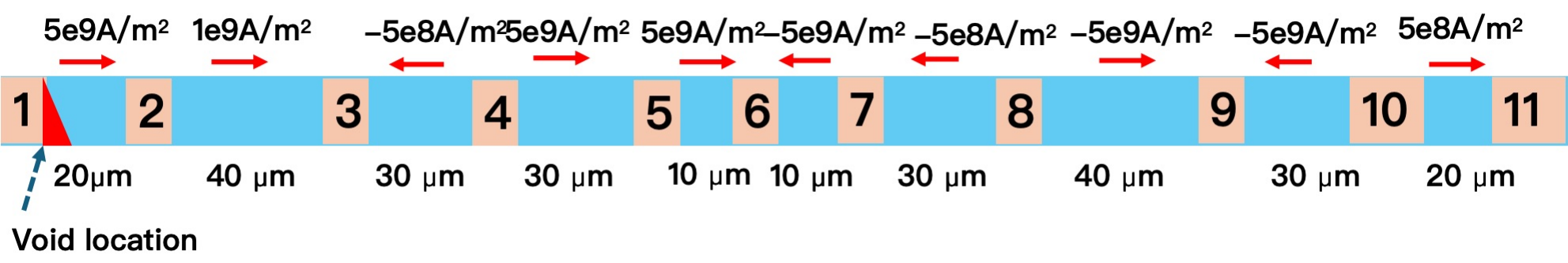}
        \label{fig:five_seg_line}
    }
    \\
    \vspace{-8pt}
    \hspace{-15pt}%
    \subfloat[Node \#1 (node with void) in (a).]{
        \includegraphics[width=0.5\linewidth]{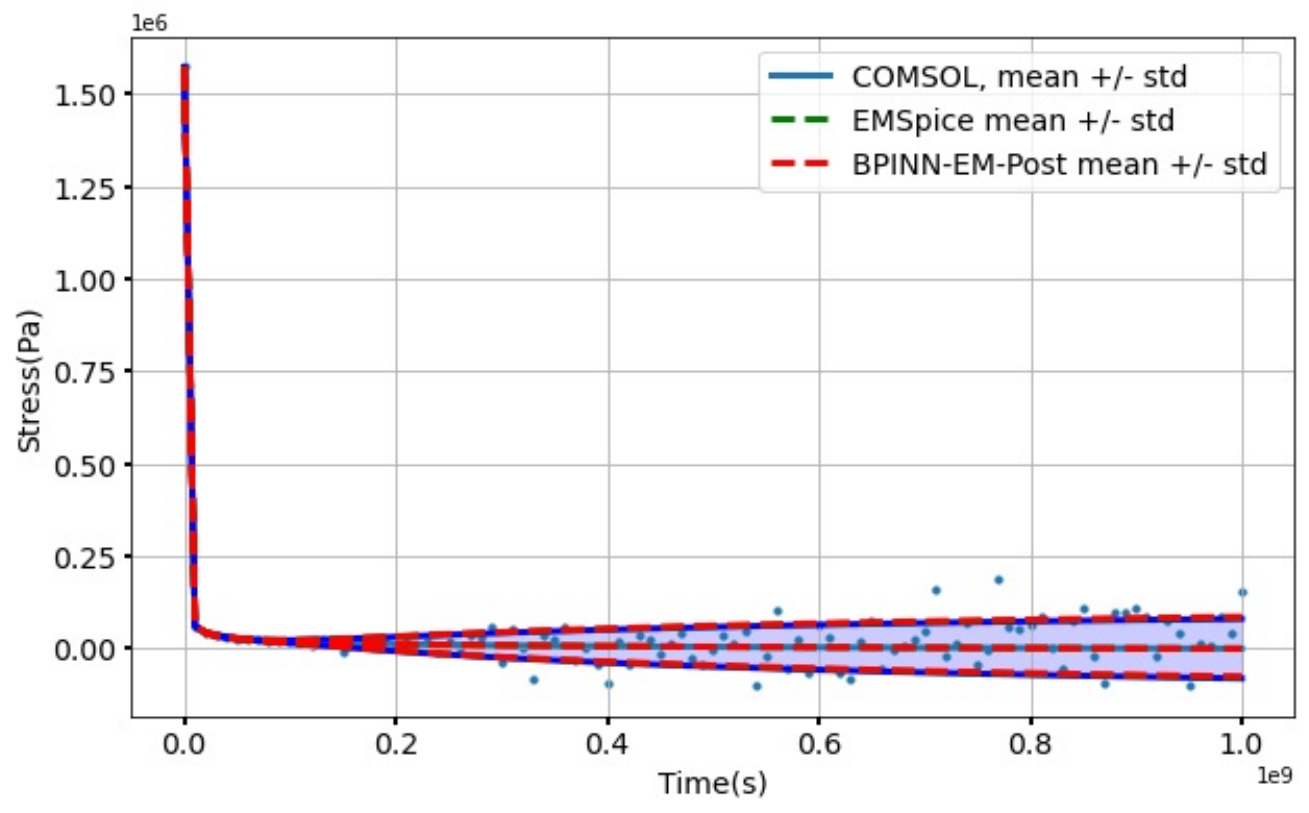}\hspace{-5pt}%
        \label{fig:result_comp_node1}
    }
    \hfill
    \subfloat[Inter-segment junction \#2 in (a).]{
        \includegraphics[width=0.5\linewidth]{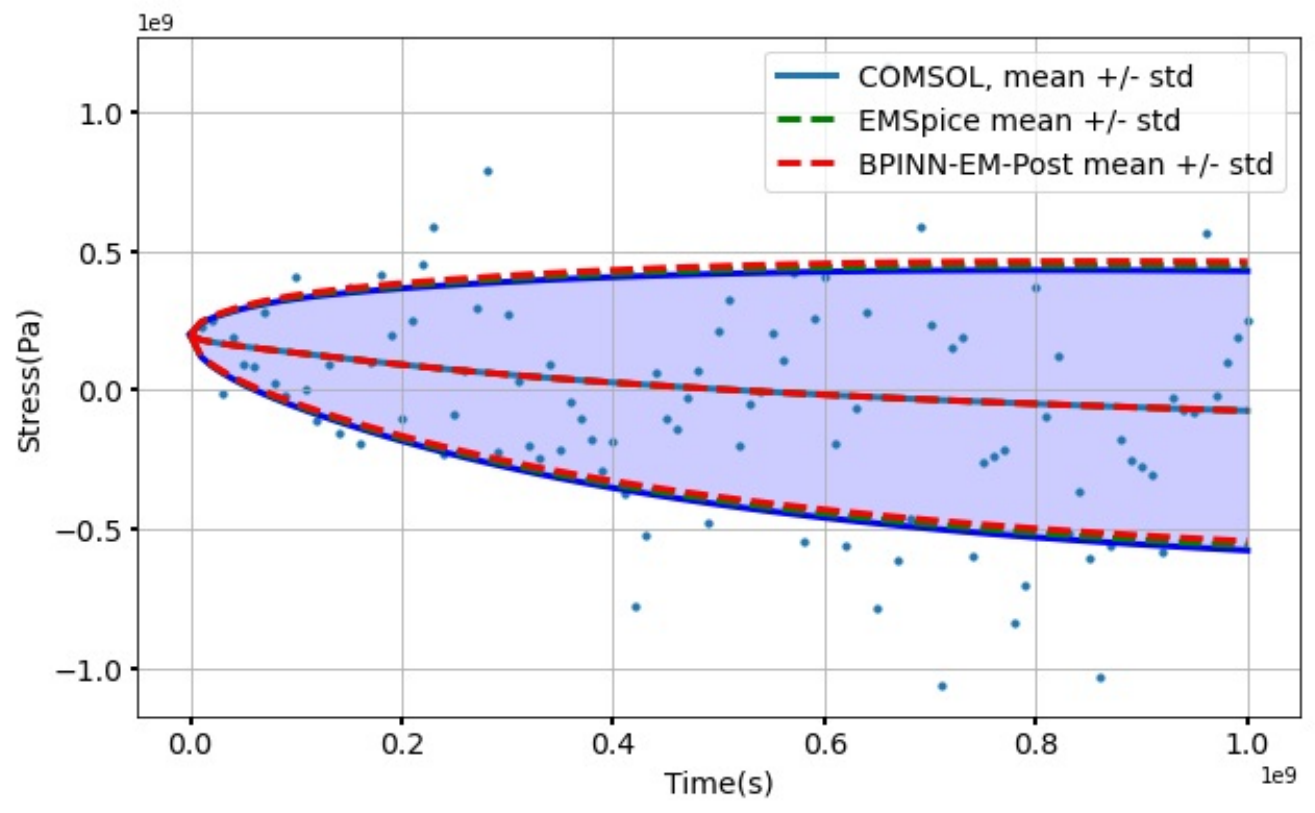}\hspace{-10pt}%
        \label{fig:result_comp_node2}
    }
    \\
    \vspace{-8pt}
    \hspace{-15pt}%
    \subfloat[Inter-segment junction \#5 in (a).]{
        \includegraphics[width=0.5\linewidth]{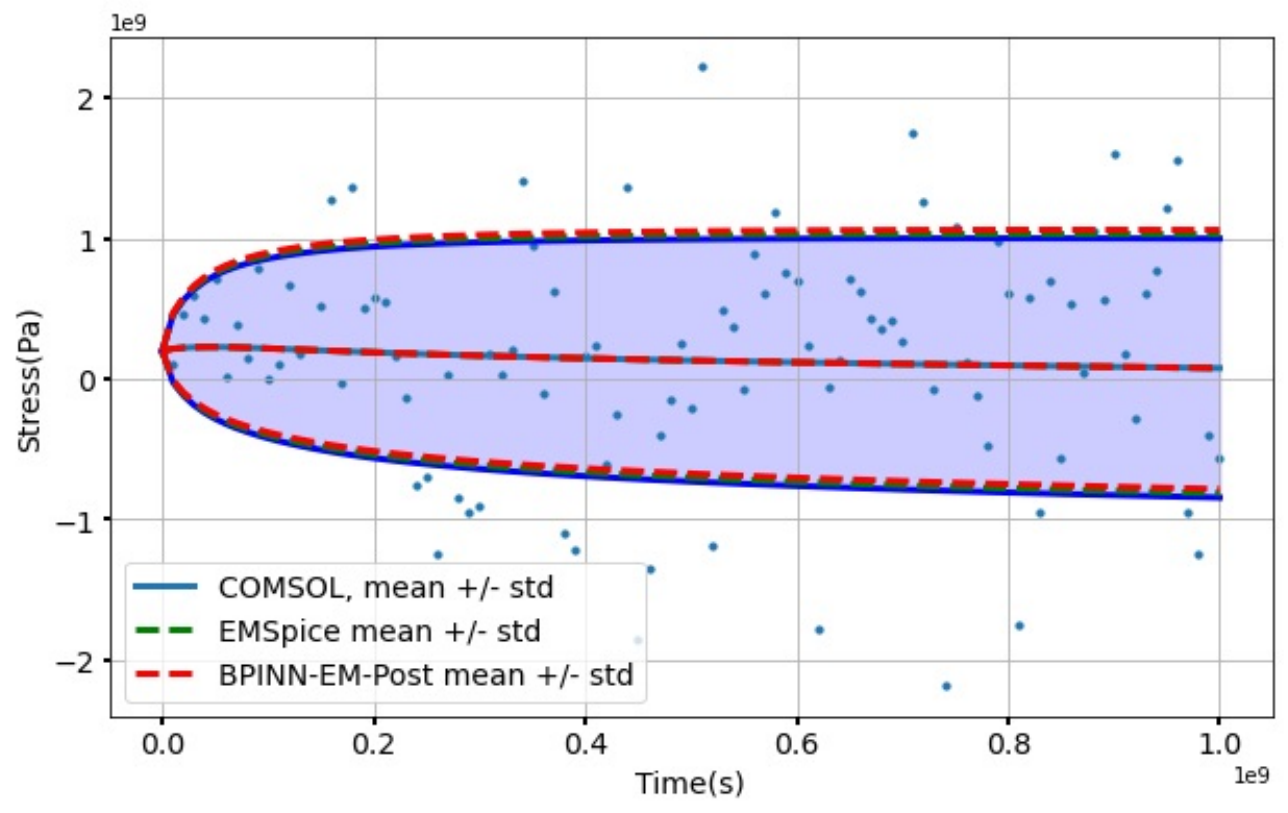}\hspace{-5pt}%
        \label{fig:result_comp_node5}
    }
    \hfill
    \subfloat[Inter-segment junction \#7 in (a).]{
        \includegraphics[width=0.5\linewidth]{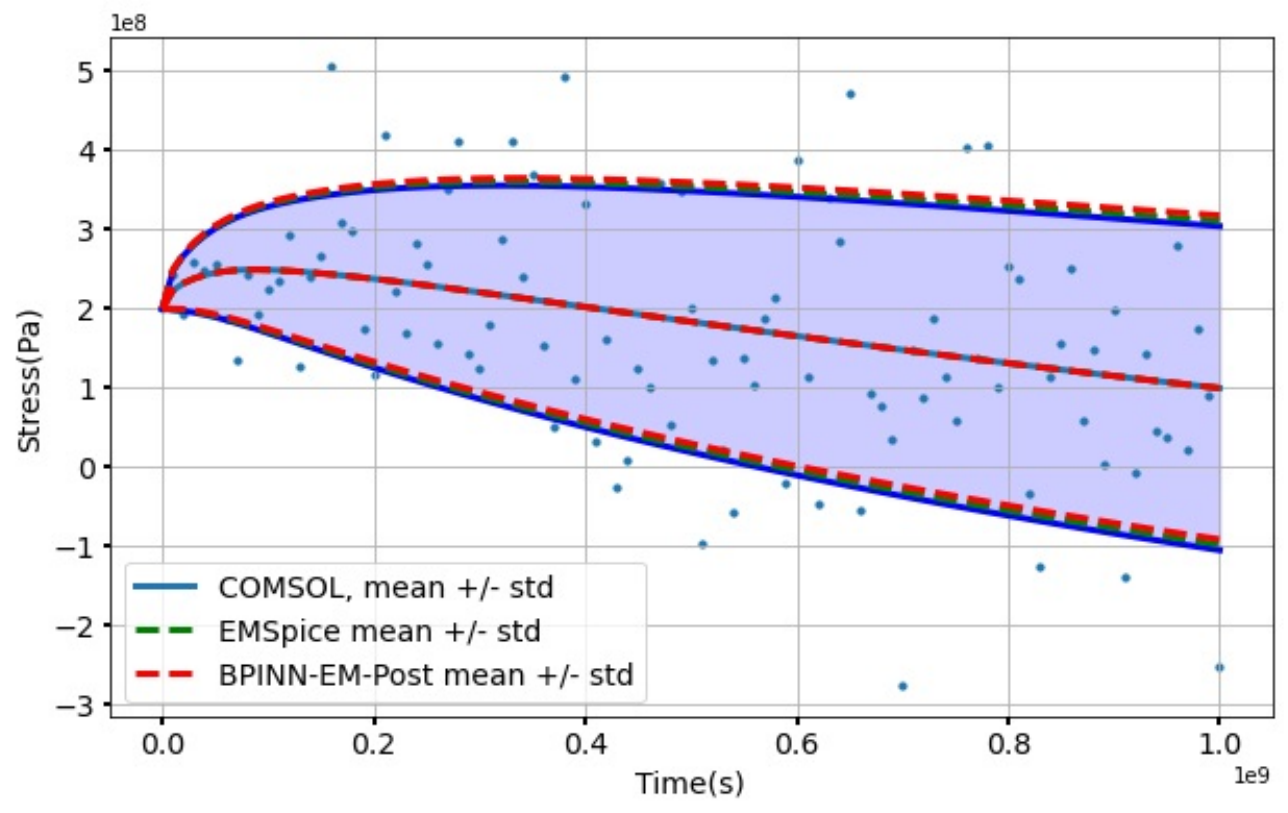}\hspace{-10pt}%
        \label{fig:result_comp_node4}
    }
\caption{\small Illustration of stress variations at junctions of a multi-segment interconnect structure. (a): Example of a ten-segment interconnect. (b), (c), (d), (e): Comparisons of variations estimation between the proposed framework, COMSOL and EMSpice at different junctions.}
\label{fig:result_comparision}
\end{figure}

To assess the efficacy and precision of our method across a variety of general multi-segment interconnect structures, we analyzed a total of 1000 randomly generated configurations, each containing between 50 and 250 segments, with each inter-segment junction connected to at most four segments. 
These structures were examined to gauge variational EM stress through the use of COMSOL~\cite{comsol:heat'2014}, EMSpice~\cite{SunYu:TDMR'20}, and our proposed \textit{\textbf{BPINN-EM-Post}}, utilizing 30 samples to measure variations in stochastic EM stress. 
We should notice that other existing works~\cite{Jin:ICCAD'22, Lamichhane:ICCAD'23} are not applicable for stochastic analysis, while~\cite{Lamichhane:ICCAD'24} cannot be employed in post-voiding phase, thus these methods are not available for numerical comparison.
The findings, detailing accuracy and performance metrics for multi-segment interconnect structures across different segment counts, are showcased in Tab.~\ref{tab:result_statistics_line}. 
Our method exhibits superior acceleration than prior arts, with error rate of only 1.10\% at most, compared with COMSOL and less than 0.2\% compared with EMSpice.
In addition, Training and Inference time shows nearly linear growth with respect to the number of wires, indicating its scalability to larger interconnect structures.

The observed range of EM stress across these structures spanned from $-4.31\times10^9$ Pa to $5.94\times 10^9$ Pa. 
Accuracy was quantified using RMSE defined as $RMSE = (RMSE_{mean} + RMSE_{std})/2$. 
For dataset comprising 1000 multi-segment interconnect structures, the average RMSE relative to COMSOL stood at $7.07\times10^7$~Pa, equating to an error margin around 0.69\%. 
In comparison, against EMSpice, the average RMSE was recorded at $9.5\times10^6$~Pa, reflecting an error rate about 0.093\%.

Fig.~\ref{fig:result_comparision} demonstrates the variations in EM stress, expressed through both mean and standard deviation, for a ten-segment interconnect structure for simplicity.
The physical and electrical characteristics of which are shown in Fig.~\ref{fig:five_seg_line}. 
The variations in EM stress at the selected junctions between segments of this interconnect structure are detailed in Fig.~\ref{fig:result_comp_node1}, Fig.~\ref{fig:result_comp_node2}, Fig.~\ref{fig:result_comp_node5}, and Fig.~\ref{fig:result_comp_node4}. 
In these comparisons, our proposed \textbf{\textit{BPINN-EM-Post}} is shown to precisely predict the variations in EM stress, aligning closely with the results obtained from COMSOL~\cite{comsol:heat'2014} and EMSpice~\cite{SunYu:TDMR'20}.

In the analysis of stochastic EM stress variances for multi-segment interconnect structures, COMSOL's average runtime for 1000 multi-segment interconnect structures is approximately 7500 seconds, while EMSpice takes about 2100 seconds in average. 
For \textit{\textbf{BPINN-EM-Post}}, variations in stochastic EM stress are estimated by sampling from the BNN via HMC. 
Prior to this, the analytical solutions must ascertain the optimal atomic flux at the termini of each segment within the structures, leveraging a PINN. 
Thus, both the sampling (training) period of the PINN and the inference duration of the BNN are accounted for in the runtime evaluation. 
For the multi-segment interconnects assessed, the average runtime for \textit{\textbf{BPINN-EM-Post}} is noted to be approximately 31 seconds, significantly accelerating the estimation process by a factor of $240\times$ faster than COMSOL and $67\times$ quicker than EMSpice. 
This efficiency marks our approach as vastly superior in terms of speed for calculating variational EM stress.
\section{Conclusion}
\label{sec:Conclusion}
In this paper, we introduced a novel machine learning-based stochastic analysis framework, termed {\it \textbf{BPINN-EM-Post}}, to efficiently estimate variational Electromigration (EM)-induced stress effects. 
This framework combines closed-form analytical solutions with a PINN to ensure stress continuity and atomic flux conservation during post-void stress calculations. 
{\it \textbf{BPINN-EM-Post}} effectively utilizes analytical solutions for individual wire segments, avoiding the need to model complex initial stress distributions. 
At the inter-segment level, physics constraints, such as stress continuity and atomic flux conservation, are enforced through the neural network, resulting in a more efficient and accelerated training process by reducing the number of variables in the loss function. 
The BPINN framework further enhances the approach by providing uncertainty quantification and enabling variational analysis, ensuring robust and efficient modeling of EM stress evolution. 
Experimental results demonstrate substantial computational improvements, achieving a $240\times$ speedup compared to FEM-based Monte Carlo simulations and a $67\times$ acceleration over FDM-based methods. 
These findings underscore the potential of the BPINN framework for scalable and accurate variational EM stress analysis.

\bibliographystyle{ieeetr}
\bibliography{ref/EM_basics, ref/mscad, ref/neural_network}

\end{document}